\documentclass{article}

\usepackage[preprint, nonatbib]{neurips_2026}

\usepackage[utf8]{inputenc} 
\usepackage[T1]{fontenc}    
\usepackage{hyperref}       
\usepackage{url}            
\usepackage{booktabs}       
\usepackage{amsfonts}       
\usepackage{nicefrac}       
\usepackage{microtype}      
\usepackage{xcolor}         

\usepackage{graphicx}
\usepackage{booktabs}
\usepackage{multirow}
\usepackage{xspace}
\usepackage{amsmath}
\usepackage{wrapfig}
\usepackage{mathtools}

\hypersetup{
colorlinks,
linkcolor=,
urlcolor=magenta,
}

\usepackage[table]{xcolor}
\usepackage{makecell}
\usepackage{arydshln}
\usepackage{multirow}
\usepackage{graphicx}
\usepackage{capt-of} 
\usepackage{bm}
\definecolor{colorfirst}{RGB}{255,153,153}
\definecolor{colorsecond}{RGB}{255,204,153}
\definecolor{colorthird}{RGB}{255,255,153}

\usepackage{fontawesome5}

\makeatletter
\renewcommand\paragraph{\@startsection{paragraph}{4}{\z@}{0.7ex plus 0.2ex minus 0.1ex}{-1em}{\normalfont\normalsize\bfseries}}
\makeatother

  {\par} 

\usepackage{enumitem}
\setlist{nosep} 

\newcommand{\method}{\textbf{StreetNVS}\xspace}
\title{Effective Multi-sensor Conditioning for Street-view Novel-view Synthesis}

\author{Zhengfei Kuang$^{1}$ \hspace{10mm} 
Adam Sun$^{1}$ \hspace{10mm}  
Liyuan Zhu$^{1}$ \hspace{10mm} 
Tong Wu$^{1}$ \hspace{10mm} \\
\textbf{Shengqu Cai}$^{1}$ \hspace{10mm} 
\textbf{Jonathan Tremblay}$^{2}$ \hspace{10mm} 
\textbf{Iro Armeni}$^{1}$ \hspace{10mm} 
\textbf{Ehsan Adeli}$^{1}$ \hspace{10mm} \\
\textbf{Lior Yariv}$^{1}$ \hspace{10mm} 
\textbf{Gordon Wetzstein}$^{1}$ \hspace{10mm} \\ \\
$^{1}$Stanford Univerity \hspace{14mm} $^{2}$NVIDIA \\
}

\begin{document}

\maketitle

\begin{abstract}
Modern vehicle platforms are equipped with a rich sensor suite, including LiDAR, calibrated multi-camera rigs, and accurate ego-motion, that in principle offers strong signal for re-rendering a driving scene from novel viewpoints.
A growing line of recent work leverages video diffusion models for this task, using their generative priors to synthesize plausible novel views from sparse vehicle observations.
In practice, however, existing methods exploit only a fragment of this signal, and their quality tends to degrade as the target trajectory departs from the recorded driving path.
We argue that this is fundamentally a \textit{multi-sensor fusion problem}: sparse LiDAR reprojections supply accurate but incomplete metric geometry, surround-view reference imagery supplies dense appearance but no metric depth, and camera poses tie the two together across views.
We introduce \method, a video diffusion framework that jointly conditions on all three signals through a Reference-Enhanced Camera Attention module based on a relative ray-level positional encoding. We develop a two-stage curriculum training strategy that gradually exposes the model to increasingly sparse LiDAR.
On the Waymo Open Dataset, \method substantially outperforms state-of-the-art baselines under sparse LiDAR conditioning, matches methods that rely on $10$--$100\times$ denser point clouds. We further show capabilities of synthesizing coherent videos along extreme out-of-trajectory paths such as elevation, lane-shift, pullback, and rotation. Our website:~\url{https://streetnvs.github.io}
\end{abstract}

\section{Introduction}
\label{sec:introduction}

\looseness=-1
Reconstructing and digitally re-experiencing a driving scene from novel viewpoints is a critical capability with growing practical importance across automotive and transportation domains. High-quality street-view novel-view synthesis (NVS) underpins a wide range of downstream applications: rendering immersive user views for incident review and navigation, generating diverse training data for autonomous driving agents and world modeling, supporting driver-assistance features such as blind-spot visualization and parking guidance, and enabling realistic closed-loop simulation for safety evaluation. Unlike traditional 3D reconstruction settings, which typically operate offline on densely captured imagery with methods such as 3D Gaussian Splatting~\cite{kerbl3Dgaussians} or other neural rendering methods~\cite{tewari2022advances}, vehicle-based streetscape reconstruction is fundamentally challenging: the scene is highly dynamic and source cameras are mounted directly on the vehicle, providing only sparse and biased coverage of the surrounding environment. As a result, target viewpoints often fall well outside the envelope of observed views, demanding strong extrapolation capabilities rather than interpolation for NVS.

\begin{figure}[t!]
  \centering
  \includegraphics[width=\textwidth]{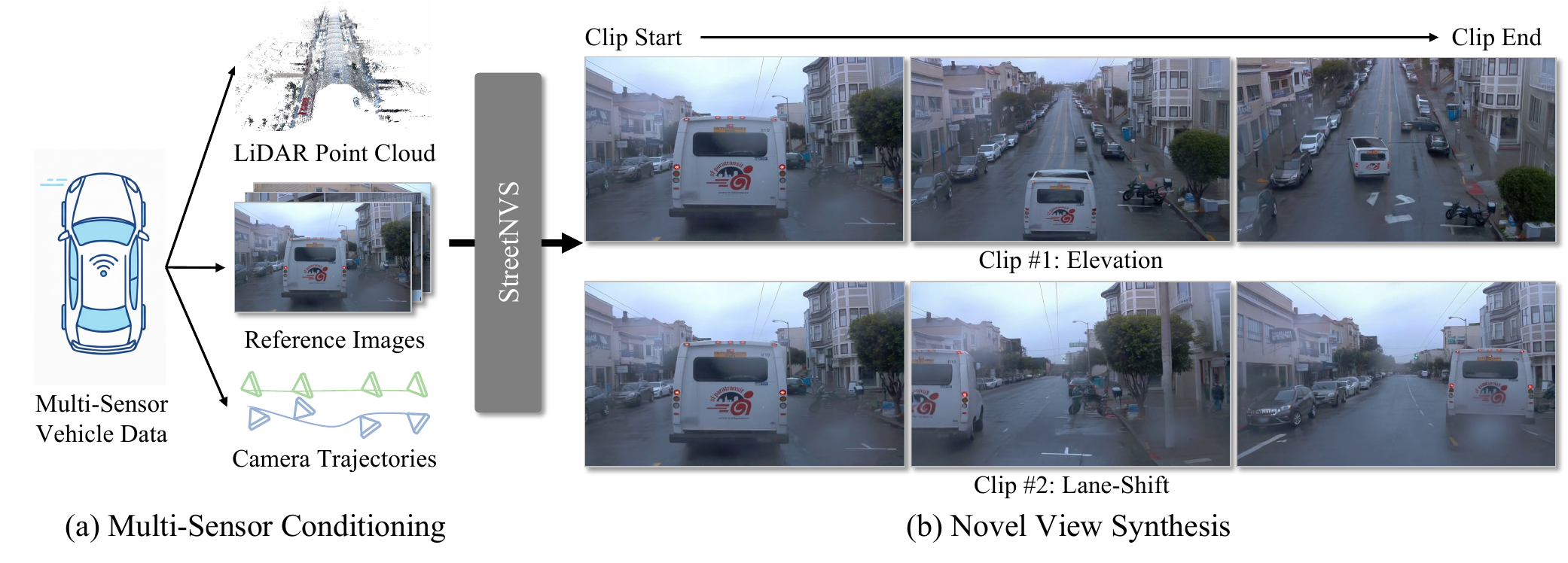}
  \caption{\textbf{\method for Street-View Novel-View Synthesis.}
\looseness=-1
\textbf{(a)} Given multi-sensor data from a vehicle rig (LiDAR point clouds, reference images, and camera trajectories), \method synthesizes street-view videos along arbitrary novel trajectories.
\textbf{(b)} Two NVS examples: an elevation trajectory lifting the camera toward a bird's-eye view (top), and a lane-shift trajectory displacing it from the original driving path (bottom).
By jointly leveraging LiDAR geometry and cross-camera references, \method produces high-fidelity views far from the vehicle's own observations.}
\label{fig:teaser}
\vspace{-15pt}
\end{figure}

To address these challenges, a growing body of recent work has turned to video diffusion models~\cite{wan2025wan, vace, xu2024camco}, which provide strong generative priors over natural driving scenes. Among these, methods such as FreeVS~\cite{wang2024freevs}, Gen3C~\cite{ren2025gen3c}, and StreetCrafter~\cite{yan2024streetcrafter} leverage dense LiDAR point clouds as a powerful geometric anchor: they reproject the LiDAR into the target novel view and feed the resulting buffer directly to the diffusion model as a conditioning signal, effectively solving an inpainting task using the video generation model. This explicit geometric grounding shown to substantially improve consistency and reduce hallucination, yielding impressive reconstruction quality when the reprojected LiDAR point cloud is dense. However, as the target view moves farther from the observed viewpoints on the vehicle, the reprojected point cloud becomes increasingly sparse, making the inpainting task ambiguous. 

\begin{wrapfigure}{r}{0.52\textwidth}
    \vspace{-15pt}
    \centering
    \includegraphics[width=0.50\textwidth]{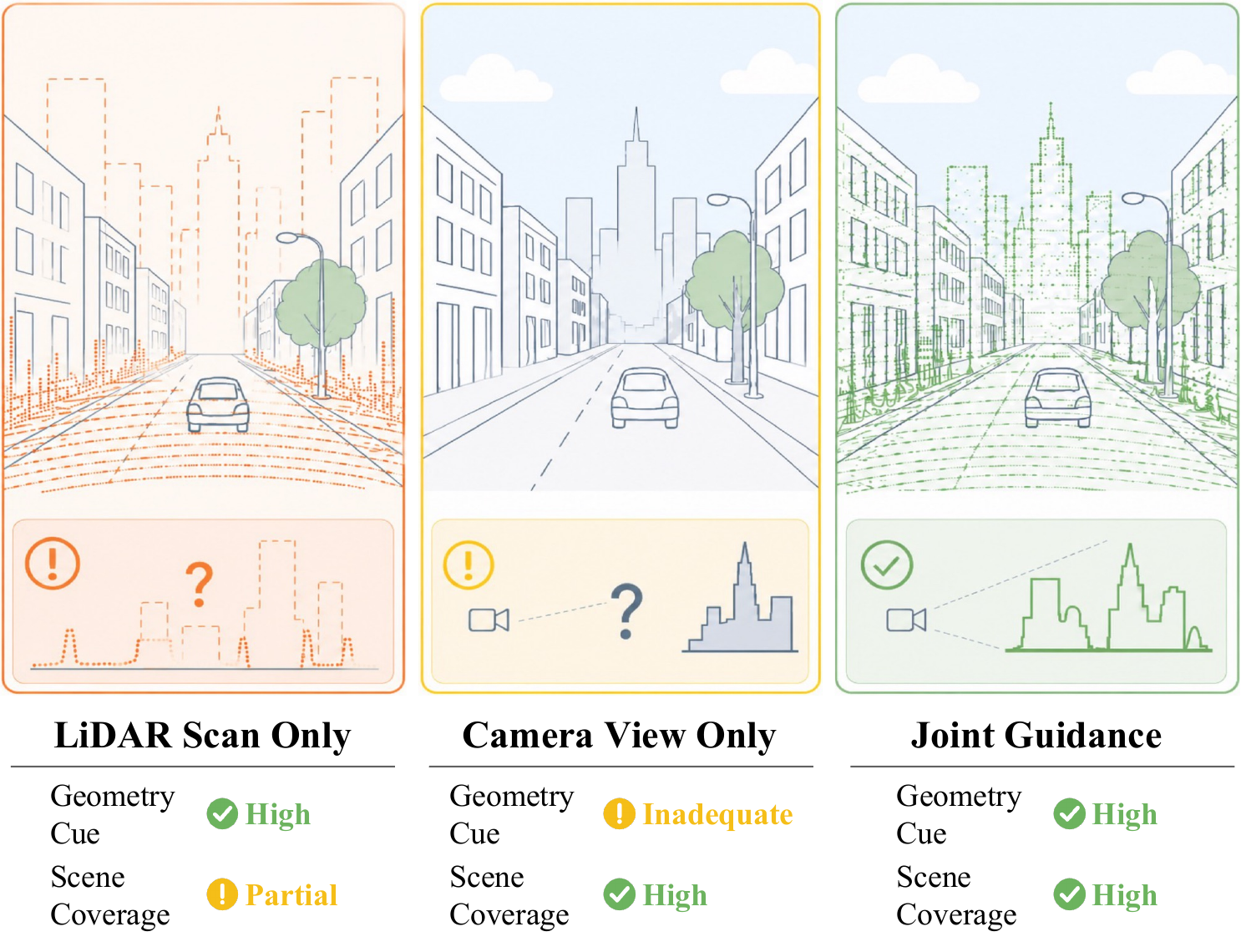}
    \vspace{-5pt}
    \caption{\textbf{Complementary Conditioning Signals.}
    LiDAR alone (left) lacks scene coverage and reference views alone (middle) lack metric geometry; jointly leveraging both (right) provides high geometric fidelity with full coverage.}
    \label{fig:hybrid}
    \vspace{-5pt}
\end{wrapfigure}

\looseness=-1
Our method, dubbed \method, is designed to optimize the quality of synthesized views far from the vehicle-mounted source camera poses where reprojected point clouds are extremely sparse.
Our key insight is that the vehicle rig also provides surround-view imagery from synchronized cameras, which complements LiDAR: as illustrated in Figure~\ref{fig:hybrid}, LiDAR provides accurate but sparse and incomplete 3D structure, while reference views offer dense appearance coverage but lack metric geometry.
Existing LiDAR-conditioned video models such as StreetCrafter~\cite{yan2024streetcrafter} sidestep this issue with an image-to-video (I2V) formulation conditioned on the first frame, which is effective in mildly extrapolated cases but can fail catastrophically under less constrained trajectories such as large rotations away from the driving direction.
In contrast, \method jointly leverages LiDAR and source camera views as conditioning signals, adopting recent advances in camera-controlled video generation using relative spatially aware positional embeddings~\cite{li2025prope, zhang2025unified, wu2026rayrope} into our multi-sensor conditioned video generator.

Trained and evaluated on the Open Waymo Dataset~\cite{Sun_2020_CVPR_waymo} using a two-stage dense-to-balanced LiDAR curriculum, we show that StreetNVS achieves state-of-the-art performance, with improved image quality, scene faithfulness, and 3D consistency. We demonstrate the robustness of StreetNVS to varying sparse point cloud conditioning, where it exhibits little-to-no quality degradation for point clouds that are 10--100$\times$ sparser than the dense LiDAR measurements taken from the perspective of the vehicle. This capability uniquely unlocks ``extreme'' NVS scenarios where the synthesized target view is far from the vehicle-mounted camera observations, as exemplified in Figure~\ref{fig:teaser} (b).

Our contributions include:
\begin{itemize}
\item We present a novel generative framework for street-view NVS that jointly conditions a video generation model on sparse LiDAR reprojections, camera poses, and reference images, substantially relaxing the dense-LiDAR requirement of prior approaches.

\item We design a camera control module that incorporates multiple reference images and target video latents, providing robust appearance anchoring across changing viewpoint without relying on dense geometric input.

\item We establish state-of-the-art robustness to sparse LiDAR conditioning and extreme novel viewpoints, matching or exceeding baselines that require roughly $10\times$ denser LiDAR, demonstrated through comprehensive experiments across various LiDAR sparsity levels.
\end{itemize}

\section{Related Work}\vspace{-3pt}
\paragraph{Street-View Novel View Synthesis.}
Early work on NVS for large-scale urban environments~\cite{zhou2024hugs, yang2023unisim} relied on Neural Radiance Fields (NeRF) to model scene geometry and dynamic objects~\cite{mildenhall2021nerf, tancik2022block, wu2023mapnerf}. More recent approaches are based on 3D Gaussian Splatting~\cite{kerbl3Dgaussians} and its driving-scene extensions~\cite{yan2024street, fischer2024dynamic, zhou2024drivinggaussian, zhao2025drivedreamer4d, chen2024omnire, zhao2025recondreamer++}, showing substantially improved rendering throughput and visual quality.
However, these methods are fundamentally reconstructive: their fidelity is tightly coupled to the density and coverage of the source views, and they struggle to extrapolate beyond the captured trajectory.
To overcome this, a growing line of work casts street-view NVS as a generative problem, leveraging diffusion-based video models conditioned on geometric priors~\cite{wang2024freevs, zhao2025recondreamer++}; StreetCrafter~\cite{yan2024streetcrafter}, for instance, trains a video generative model to synthesize novel views and then refines a Gaussian splatting representation from the generated views.
Yet their performance still degrades when the target camera pose deviates significantly from the recorded trajectory, partly because the LiDAR point cloud reprojected into the target view becomes increasingly sparse.

\paragraph{3D Reprojection-Based Conditioning.}
A growing body of work treats 3D reconstruction as a structural scaffold for diffusion-based video generation. Typically, these methods condition the diffusion model on rendered geometry via ControlNet~\cite{Zhang_2023_controlnet}, VACE-style modules~\cite{vace}, or embedding residuals added to the noised latents, framing generation as a pixel-level inpainting task. A common approach projects geometry predicted from a 3D vision model~\cite{dust3r_cvpr24, depthanything3, wang2025vggt, huang2025vipe, kuang2025buffer, liu2025scaling} and trains video diffusion models to generate missing texture in the reprojections~\cite{yu2024viewcrafter, yu2025trajectorycrafter, zhang2025advances, wu2025spmem, zhai2025stargen, ren2025gen3c, zhang2025wvd, hou2024CamTrol, zheng2025vidcraft3}. Diffusion-as-Shader~\cite{gu2025das}, MotionStream~\cite{shin2026motionstream}, and Edit-by-Track~\cite{lee2025editbytrack} use 3D tracking videos as control signals, while MosaicMem~\cite{yu2026mosaicmem} reprojects patchified tokens for memory efficiency. A related line of works~\cite{liu20243dgs_enhancer, wu2025difix3d, zhu2026gaussfusion} applies generative priors to repair artifacts and inpaint unseen regions in 3D Gaussian splats. 
While these methods establish a general-purpose paradigm for 3D-conditioned video generation, they rely on a feed-forward models predicting geometry from input images, hence can be metrically inconsistent. In this work, we focus on the automotive setting and exploit the accurate LiDAR and multi-view imagery directly available from the vehicle rig.

\paragraph{Camera Pose--Controlled Video Generation. } 
Controlling camera motion in pretrained video generation models has recently emerged as a crucial problem, enabling controllable scene exploration and serving as a building block for world models.
One line of work encodes camera parameters, such as extrinsic matrices~\cite{wu2024cat4d, bai2025recammaster, wang2024motionctrl} or Plücker ray maps~\cite{bahmani2025ac3d, he2025cameractrl, he2025cameractrl2, xu2024camco, zheng2024cami2v, li2025realcam, kuang2024cvd}, and fuses them into the latent features of pretrained video models.
These approaches are lightweight but resemble absolute positional encoding, potentially limiting generalization to unseen motions.
More recent works introduce relative camera-aware positional encodings~\cite{xiong2023cape, miyato2023gta, li2025prope, zhang2025unified} into DiT-style architectures for improved 3D consistency, with RayRoPE~\cite{wu2026rayrope} further injecting depth-based 3D correspondences.
Differently from these works, we introduce LiDAR measurements to resolve scale ambiguities and further incorporate multiple reference views via cross-attention with camera-aware positional encodings, achieving significant gains in 3D reconstruction quality on street-view NVS.

\label{sec:related}

\section{Method}
\label{sec:method}

We consider autonomous driving scenarios in street-view environments within a temporal window of $F$ frames.
Modern vehicle platforms are equipped with multiple synchronized sensors, including $N$ surround-view RGB cameras with fixed extrinsics relative to the vehicle, producing frames $\{\bm{v}_i^{(1,\dots,N)}\}_{i=1}^{F}$ with associated camera poses $\{\bm{T}_i^{(1,\dots,N)}\}_{i=1}^{F}$, and a LiDAR sensor providing per-frame point cloud observations $\{\mathcal{L}_i\}_{i=1}^{F}$.
Given the surround-view observations of the starting frame $\bm{v}^{\mathrm{ref}} \ \dot= \ \bm{v}_1^{(1,\dots,N)}$ together with an aggregated and optionally subsampled point cloud $\mathcal{L}^{\mathrm{agg}} \subseteq \bigcup_{i=1}^{F} \mathcal{L}_i$ from the LiDAR scan, our goal is to synthesize a novel-view video $\{\bm{v}^{\mathrm{tgt}}_i\}_{i=1}^{F}$ along a target camera trajectory $\{\bm{T}^{\mathrm{tgt}}_i\}_{i=1}^{F} \in \mathrm{SE}(3)^F$ that may differ substantially from the original sensor viewpoints (e.g., elevation or third-person views).
To simulate the wide range of LiDAR densities encountered in practice, where reprojected point clouds become increasingly sparse as the target trajectory departs from the source cameras, we deliberately train and evaluate \method across multiple subsampling ratios of $\mathcal{L}^{\mathrm{agg}}$.
The synthesized video should
(1) remain faithful to the accumulated RGB and LiDAR observations,
(2) closely follow the prescribed target trajectory, and
(3) preserve correct scene geometry and dynamic road behaviors.

\subsection{Preliminaries}

To address this challenging setting, we build our pipeline on top of a state-of-the-art large video diffusion model~\cite{wan2025wan}, leveraging the strong appearance prior learned from massive video pretraining.
For controllable generation, our framework further builds on recent advances in relative camera pose encodings, which inject geometric structure directly into the transformer's attention layers.

\paragraph{\textbf{Video Diffusion Models.}}
Given user conditions $c$ such as a text prompt, input frames, or camera poses, video diffusion models generate a video that follows the conditioning signal through an iterative denoising process. We adopt the Flow Matching formulation~\cite{lipman2022flow}, which defines a probability path by linearly interpolating clean data $\bm{z}_0 \sim P_{\mathrm{data}}$ and Gaussian noise $\bm{z}_1 \sim \mathcal{N}(\bm{0}, \bm{I})$ as $\bm{z}_t = t\bm{z}_1 + (1-t)\bm{z}_0$ for $t \in [0, 1]$, with associated velocity $\bm{z}_1 - \bm{z}_0$. A network $\bm{u}_\theta(\bm{z}_t, t, c)$ is trained to predict this velocity by minimizing the standard Flow Matching MSE loss:
\begin{equation}
    \mathcal{L}_{\mathrm{FM}} =
    \mathbb{E}_{t, c, \bm{z}_0, \bm{z}_1}
    \left\lVert \bm{u}_\theta(\bm{z}_t, t, c) - (\bm{z}_1 - \bm{z}_0) \right\rVert^2.
\end{equation}
At inference, samples are generated by integrating the learned ODE $\dot{\bm{z}_t} = \bm{u}_\theta(\bm{z}_t, t, c)$ from $t = 1$ to $t = 0$ in discrete steps, starting from pure Gaussian noise.


\paragraph{\textbf{Relative Camera Pose Encodings.}}
Transformer attention computes outputs as a weighted aggregation of value vectors $\bm{v}$, with weights given by the softmax of inner products between queries $\bm{q}$ and keys $\bm{k}$, all linearly projected from token features. Relative camera pose encoding injects geometric structure into this computation by associating each token with a per-token transformation $\bm{T} \in \mathrm{SE}(3)$ derived from the camera pose, and modifying $\bm{q}$, $\bm{k}$, and $\bm{v}$ through a block-diagonal matrix $\bm{D} = \bm{I}_{d/4} \otimes \bm{T}$ that repeats $\bm{T}$ along the diagonal to span the $d$ feature channels~\cite{kong2024eschernet, miyato2023gta}:
\begin{equation}
    \mathrm{Attn}^{\mathrm{relcam}}(\bm{q}, \bm{k}, \bm{v}) =
    \bm{D}^\top \mathrm{Attn}\!\left(\bm{D}^\top \bm{q},\, \bm{D}^{-1} \bm{k},\, \bm{D}^{-1} \bm{v}\right).
    \label{eq:rel-attn}
\end{equation}\vspace{-2pt}
\looseness=-1
Under this formulation, the attention score $A_{mn} \propto \bm{q}_m^\top \bm{D}_m \bm{D}_n^{-1} \bm{k}_n$ depends only on the relative transformation between token pairs, where $m, n$ are token indices, providing explicit geometric awareness across views. We build on UCPE~\cite{zhang2025unified}, which applies this idea at the ray level: rather than sharing a single $\bm{T}$ across all tokens of a frame, each token is assigned its own viewing ray, and a ray-to-world transformation $\bm{T}_m = \begin{bsmallmatrix} \bm{R}_m & \bm{t}_m \\ \bm{0}^\top & 1 \end{bsmallmatrix}$ is applied to its $\bm{q}$, $\bm{k}$, $\bm{v}$, where $\bm{R}_m = [\bm{x}_m, \bm{y}_m, \bm{z}_m]$ defines a local ray frame and $\bm{t}_m$ is the camera center.

\subsection{Overview of \method} 
\begin{figure*}[t]
  \centering
  \includegraphics[width=\textwidth]{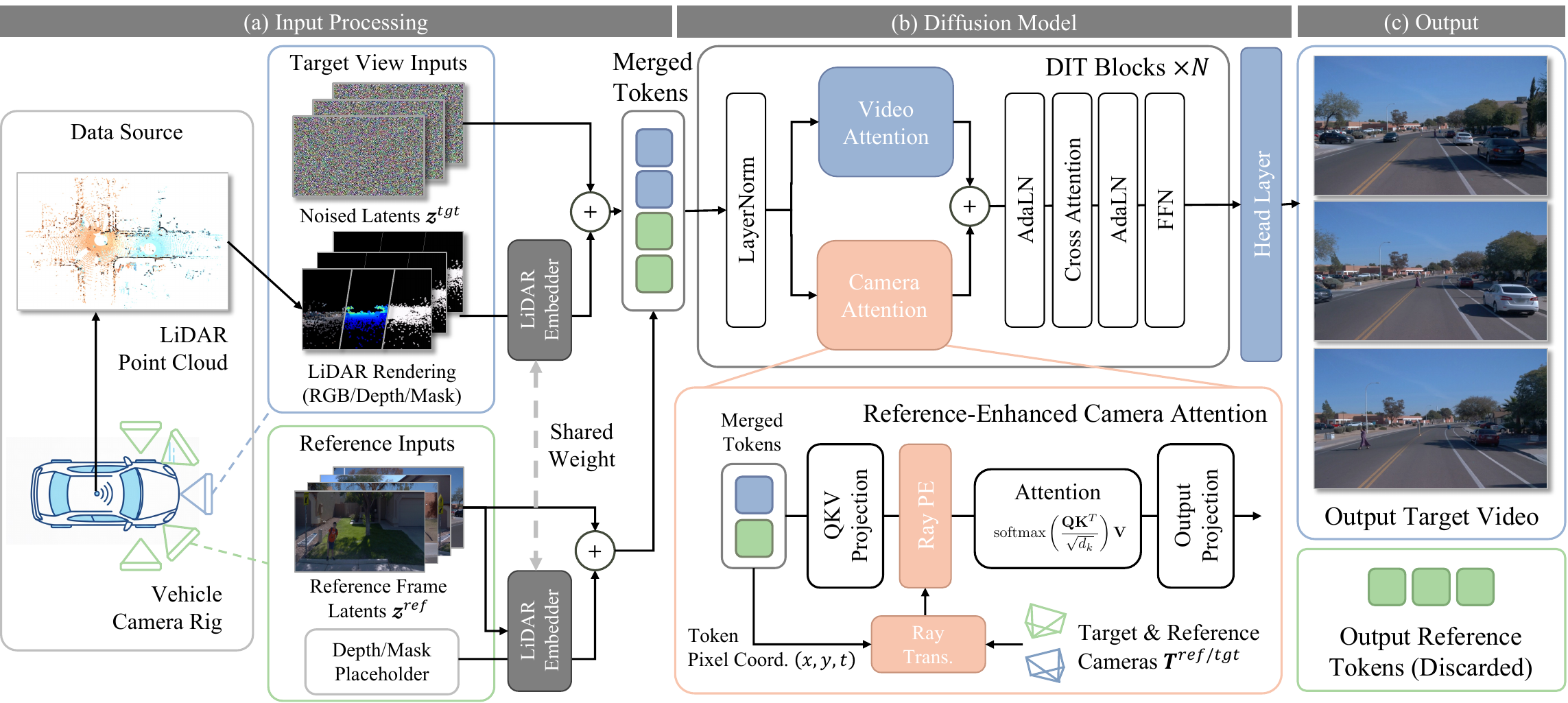}
  \looseness=-1
  \caption{\textbf{Overview of \method.}
    Our framework performs street-view novel-view synthesis by conditioning a Diffusion Transformer (DiT) on LiDAR measurements, multi-view reference imagery and camera poses.
    \textbf{(a)} The \textbf{LiDAR Embedder} extracts features from  LiDAR reprojections and merges them with the noised target latent $\bm{z}^{\textit{tgt}}$; reference latents $\bm{z}^{\textit{ref}}$ are processed by the same weight-shared embedder with depth/mask placeholders.
    \textbf{(b)} Within each DiT block, our \textbf{Reference-Enhanced Camera Attention} branch runs in parallel with the video attention, applying a ray-level positional encoding (Ray PE) over target and reference poses to align features across views.
    \textbf{(c)} The target tokens are decoded into the \textbf{synthesized novel-view video}, while reference tokens are discarded.}
  \label{fig:pipeline}
\end{figure*}

\vspace{-3pt}
\looseness=-1
As motivated in Section~\ref{sec:introduction}, prior work such as StreetCrafter~\cite{yan2024streetcrafter} conditions on only a subset of the complementary signals available from the vehicle rig.
\method instead jointly leverages all geometric cues---camera poses, LiDAR, and surround-view reference imagery---in a unified Diffusion Transformer (DiT) framework, illustrated in Figure~\ref{fig:pipeline}.
We first render the aggregated LiDAR point cloud into the target trajectory, producing pixel-aligned RGB, validity mask, and normalized depth videos in the target view.
These are encoded by the video diffusion model's VAE encoder $\mathrm{Enc}(\cdot)$. The resulting latents are concatenated channel-wise, passed through a lightweight LiDAR embedder for alignment, and added token-wise to the noisy target latent $\bm{z}^{\mathrm{tgt}}$.
In parallel, reference views $\bm{z}^{\mathrm{ref}} = \mathrm{Enc}(\bm{v}^{\mathrm{ref}})$ from the vehicle cameras are encoded by the same VAE and passed through the \emph{same weight-shared} embedder.
The reference image occupies the RGB channel while the LiDAR-specific channels are filled with placeholders (an all-ones mask and an all-zeros depth map), so that reference and target tokens share an identical channel layout and embedding pathway.
The reference latents are then concatenated with the LiDAR-injected target latents to form the merged token sequence fed into the diffusion backbone.
 
The diffusion backbone consists of a stack of DiT blocks followed by a head layer that decodes the target tokens into the velocity prediction, while reference tokens are discarded at the output.
To enable cross-view conditioning between target and reference frames while respecting their camera poses, we augment each DiT block with a dedicated camera-control branch that injects the poses of all frames into the attention computation.
The original self-attention module and remaining layers are kept intact and jointly fine-tuned with the new branch.

\subsection{Reference-Enhanced Camera Attention}

The camera-control branch operates on the merged token sequence $\bm{z}_{1,\dots,M} = \{\bm{z}^{\mathrm{tgt}}, \bm{z}^{\mathrm{ref}}\}$ of length $M$, together with per-frame camera poses $\bm{T}^{\mathrm{view}} = \{\bm{T}^{\mathrm{tgt}}, \bm{T}^{\mathrm{ref}}\}$ covering both the target trajectory and the reference views, and per-token frame indices $\tau_{1,\dots,M}$ that map each token to its corresponding pose in $\bm{T}^{\mathrm{view}}$.
We formulate the branch as global self-attention across all tokens, with relative camera-pose positional encoding applied to queries, keys, and values.
Each token is first projected to query, key, and value:
\begin{equation}
    \bm{q} = f_q(\bm{z}), \quad
    \bm{k} = f_k(\bm{z}), \quad
    \bm{v} = f_v(\bm{z}),
\end{equation}
and then transformed by a relative camera positional encoding $\mathcal{E}$:
\begin{equation}
    \tilde{\bm{q}} = \mathcal{E}(\bm{q},\, \bm{T},\, \tau), \quad
    \tilde{\bm{k}} = \mathcal{E}^{-1}(\bm{k},\, \bm{T},\, \tau), \quad
    \tilde{\bm{v}} = \mathcal{E}^{-1}(\bm{v},\, \bm{T},\, \tau),
\end{equation}
where $\mathcal{E}^{-1}(\cdot, \bm{T}, \tau) = \mathcal{E}(\cdot, \bm{T}^{-1}, -\tau)$.
The encoding $\mathcal{E}$ has two components.
The first follows UCPE~\cite{zhang2025unified} and assigns each token a ray-level transformation $\bm{T}_m = \mathrm{RayTrans}(\bm{T}^{\mathrm{view}}, \tau_m, x_m, y_m)$ derived from the camera pose at frame $\tau_m$ and the token's image coordinates $(x_m, y_m)$.
The second injects the frame index $\tau_m$ via standard RoPE~\cite{su2024roformer}.
The two encodings are concatenated channel-wise; we do not additionally encode pixel coordinates within the image plane, since the ray-level pose encoding already captures the corresponding geometry.

\looseness=-1
The encoded queries, keys, and values are passed through self-attention; the attention output is then re-encoded with $\mathcal{E}$ and projected via the output layer $f_o$ before being passed to the feed-forward network:
\begin{equation}
    \bm{o} = f_o\!\left(
        \mathcal{E}\!\left(
            \mathrm{Attn}(\tilde{\bm{q}}, \tilde{\bm{k}}, \tilde{\bm{v}}),\,
            \bm{T},\, \tau
        \right)
    \right).
\end{equation}

\subsection{Two-Stage Progressive Training Curriculum}\vspace{-3pt}

When the full pipeline is trained jointly from the start, we empirically observe that the model struggles to follow the LiDAR rendering guidance, especially in the sparse regime.
We attribute this to two factors: (1) our backbone is a generic video diffusion model with no prior exposure to LiDAR-conditioned street-view synthesis, and (2) under high subsampling ratios, the LiDAR input is too sparse to provide a useful learning signal on its own.

\looseness=-1
To address this, we adopt a two-stage training curriculum.
In the first stage, we disable the camera-control branch and the reference-frame inputs, training the model with only the target LiDAR renderings as guidance and emphasizing denser LiDAR samples to give the model a strong initial signal.
In the second stage, we enable all components and fine-tune the full pipeline jointly, using a balanced sparsity distribution that exposes the model to the full sparsity range encountered at test time.

\section{Experiments}\vspace{-3pt}
\label{sec:experiments}

\begin{figure*}[t]
\centering
\includegraphics[width=\linewidth]{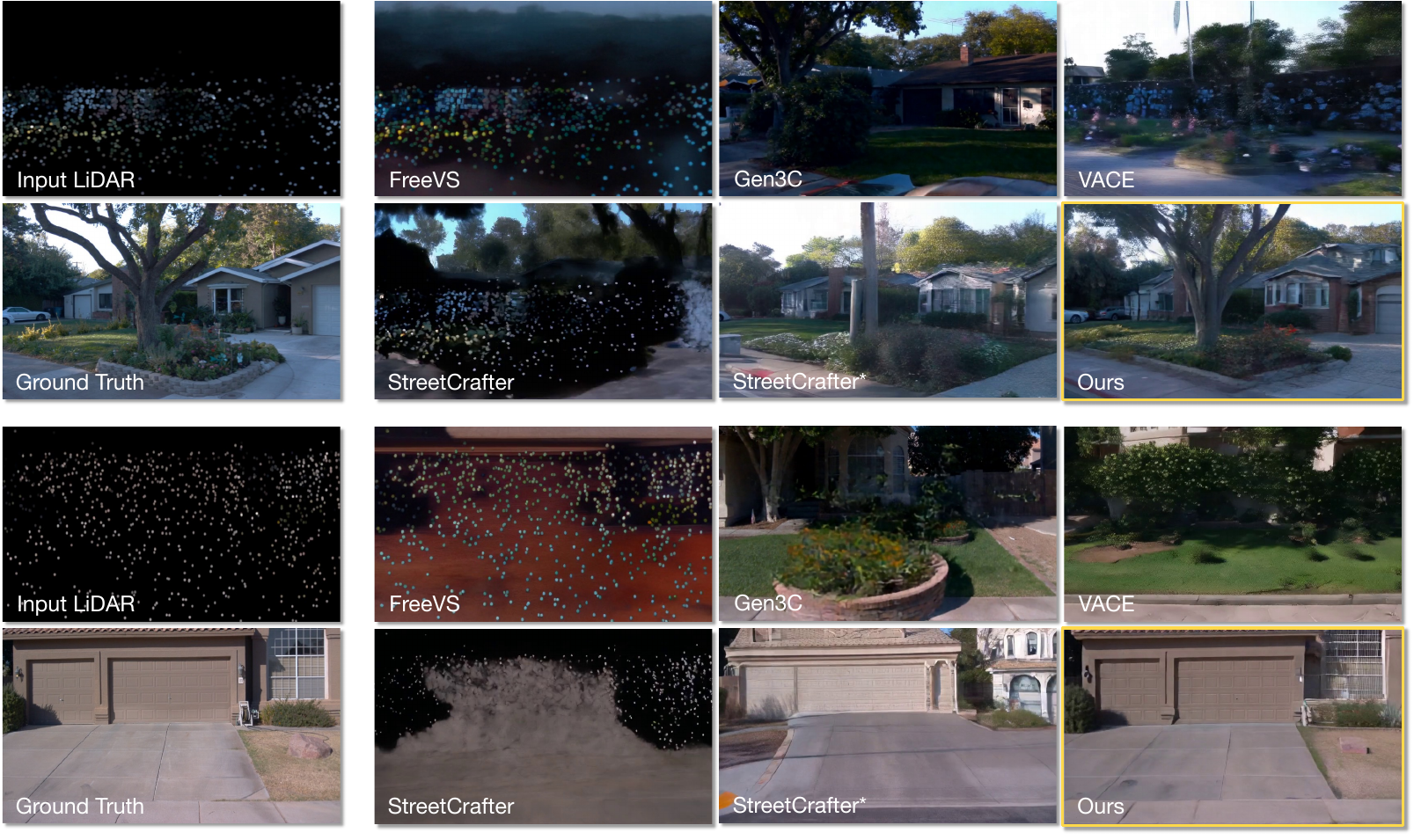}
\caption{
    \textbf{Qualitative comparison with baseline methods.} Baselines degrade substantially under sparse LiDAR conditioning. Our fine-tuned StreetCrafter (StreetCrafter*) recovers the scene only partially, with visible inconsistencies relative to the ground truth, while our model reconstructs the scene with the highest fidelity.
}
\label{fig:qualitative_comparison}
\end{figure*}

\subsection{Implementation Details}\vspace{-3pt}
We train and evaluate our model on the Waymo Open Dataset~\cite{Sun_2020_CVPR_waymo}, a large-scale autonomous driving benchmark captured by vehicles carrying five cameras and a LiDAR scanner.
For each sample, we designate one camera as the target view over a 49-frame clip and use the initial frame of the remaining four as reference views.
We use the official train and test splits.
To cover a wide range of LiDAR densities, we aggregate LiDAR scans over every 10 frames per clip and subsample the aggregated points at ratios from $0.001$ to $1$ at order-of-magnitude intervals, providing validity masks and normalized depth maps as auxiliary inputs.
\looseness=-1

Our model is built on WAN-2.2-I2V-5B~\cite{wan2025wan} and trained within the DiffSynth~\cite{diffsynth-engine2025} framework using AdamW~\cite{adamw}.
A 500-iteration warm-up ramps the learning rate from $5\times10^{-6}$ to $5\times10^{-5}$, followed by linear decay back to $5\times10^{-6}$.
Each stage is trained for 10K iterations on 8 NVIDIA H100 GPUs, taking approximately 30 hours in total.

\subsection{Novel-View Synthesis Evaluation}

We curate 402 evaluation clips from the Waymo Open Dataset test split, drawn from all five cameras and processed using the same protocol as the training set, and evaluate every model under sparse LiDAR conditioning at a sampling ratio of $0.01$.
We report PSNR, SSIM~\cite{ssim}, LPIPS~\cite{LPIPS}, and FVD~\cite{FVD}.
For baselines, we select three state-of-the-art street-view synthesis methods, namely StreetCrafter~\cite{yan2024streetcrafter}, FreeVS~\cite{wang2024freevs}, and Gen3C~\cite{ren2025gen3c}, all trained on Waymo with dense LiDAR, as well as a general-purpose video-to-video model that supports point reprojection inputs, WAN-2.1-VACE-14B~\cite{vace}.
To ensure a fair comparison, we additionally re-implement StreetCrafter on top of our backbone and retrain it across the same range of LiDAR sparsity levels, denoted as StreetCrafter*.

As shown in Table~\ref{tab:quantitative_comparison}, our method substantially outperforms all baselines across all metrics.
Qualitatively, Figure~\ref{fig:qualitative_comparison} shows that FreeVS, Gen3C, and the original StreetCrafter degrade severely under sparse LiDAR, either reproducing the scattered points directly, hallucinating incorrect scenes, or producing blurred output.
Our retrained StreetCrafter*, the strongest comparison as it is exposed to the same sparsity range as our model, recovers more of the scene but still exhibits visible inconsistencies in geometry and object placement.
In contrast, our model reconstructs the scene with the highest fidelity, faithfully reproducing both global layout and fine appearance details.

\begin{table}[t]
\centering
\small
\begin{tabular}{l  cccc}
\toprule
Method  & PSNR $\uparrow$ & SSIM $\uparrow$ & LPIPS $\downarrow$  & FVD $\downarrow$  \\
\midrule\midrule

FreeVS\cite{wang2024freevs}  & 13.19 & 0.407 & 0.721 & 171.42 \\
WAN VACE 14B\cite{vace} & 14.67 & 0.430 & 0.623  & 60.68  \\
Gen3C\cite{ren2025gen3c} & 14.49 & 0.427 & 0.602 & 29.78  \\
StreetCrafter\cite{yan2024streetcrafter} & 16.81 & 0.492 & 0.494  & 40.78  \\
StreetCrafter* & 19.22 & 0.582 & 0.377  & 12.29 \\
Ours & \textbf{20.82} & \textbf{0.619} & \textbf{0.311}  & \textbf{9.12}  \\
\bottomrule\bottomrule
\end{tabular}

\vspace{7pt}
\caption{
\textbf{Quantitative Comparisons on Street View Synthesis} at LiDAR sparsity ratio 0.01. We additionally fine-tune StreetCrafter under the same setup for fair comparison (StreetCrafter*). }

\vspace{-15pt}
\label{tab:quantitative_comparison}
\end{table}

\subsection{Ablation Study}

\begin{figure*}[t]
\centering
\includegraphics[width=\linewidth]{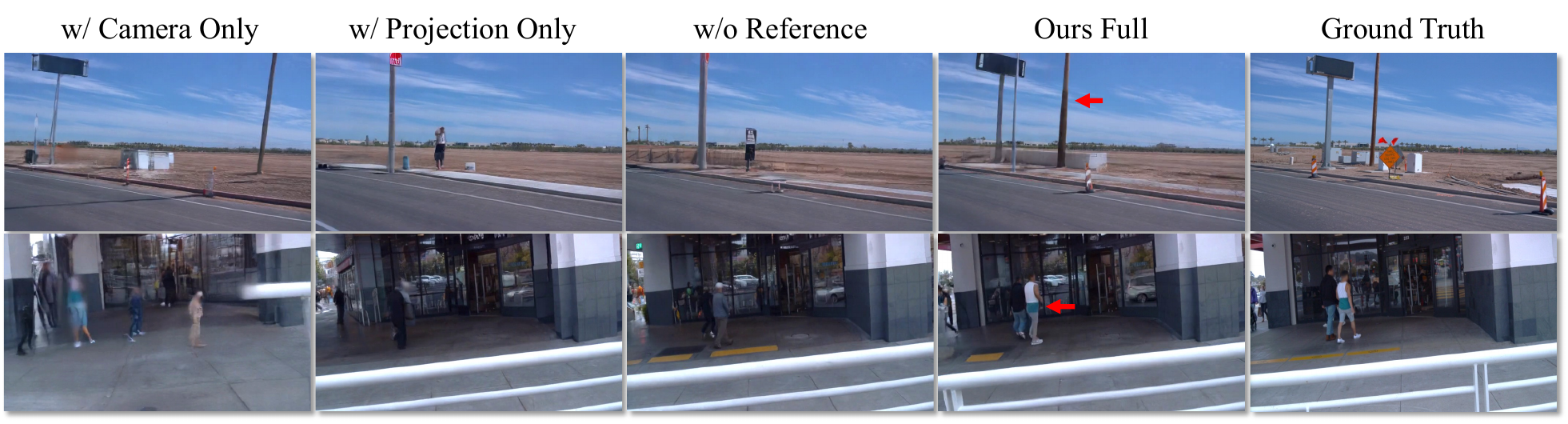}
\caption{
\textbf{Qualitative Ablation Study.} Without LiDAR projection, the model partially recovers scene identity but fails on geometry. Removing camera attention or reference views causes inconsistent generation. Our full model successfully reconstructs consistent content at the correct location.
}
\label{fig:qualitative_ablation}
\end{figure*}
\begin{table}[t]
\centering
\small
\begin{tabular}{l cccc}
\toprule
Method  
 & PSNR $\uparrow$ & SSIM $\uparrow$ & LPIPS $\downarrow$ & FVD $\downarrow$ \\
\midrule\midrule
Ours w/ Camera Only & 17.60 & 0.541 & 0.452 & 15.77 \\
Ours w/ Projection Only & 19.98 & 0.598 & 0.354 & 12.00 \\
Ours w/o Reference Views & 20.34 & 0.607 & 0.334 & 10.28 \\
Ours w/o Progressive Training & 19.51 & 0.583 & 0.365 & 11.13 \\
Ours Full & \textbf{20.82} & \textbf{0.619} & \textbf{0.311} & \textbf{9.12} \\

\midrule
\bottomrule
\end{tabular}

\vspace{5pt}
\caption{\textbf{Ablation Study at LiDAR sparsity ratio 0.01.} Removing any of the LiDAR projection embedding, camera attention, reference views, or two-stage curriculum degrades performance, with the full model performing best across all metrics.}
\label{tab:ablation_study}
\end{table}

To verify the effectiveness of each component, we ablate four variants under the same evaluation protocol: \textit{Ours w/ Camera Only} relies solely on camera attention and discards LiDAR-derived inputs; \textit{Ours w/ Projection Only} uses only LiDAR reprojection videos, without camera attention or reference frames; \textit{Ours w/o Reference Views} combines LiDAR projections with camera attention but omits reference views; and \textit{Ours w/o Progressive Training} trains end-to-end in a single stage.
Table~\ref{tab:ablation_study} shows that our full model consistently outperforms all variants, with two notable findings: \textit{Ours w/o Reference Views} surpasses \textit{Ours w/ Projection Only}, indicating that camera attention alone provides a meaningful boost; and omitting the two-stage curriculum substantially degrades performance.

Qualitative results in Figure~\ref{fig:qualitative_ablation} further illustrate these trends: without LiDAR, the model partially recovers scene identity from cross-view cues but produces incorrect geometry; without camera attention or reference views, it captures consistent layout but fails to fill regions uncovered by LiDAR (e.g., the top of the pillar in the first example).
Only the full model handles both aspects, yielding results that most closely match the ground truth.

\subsection{Comparison Across LiDAR Densities}

\begin{figure*}[t]
\centering
\includegraphics[width=\linewidth]{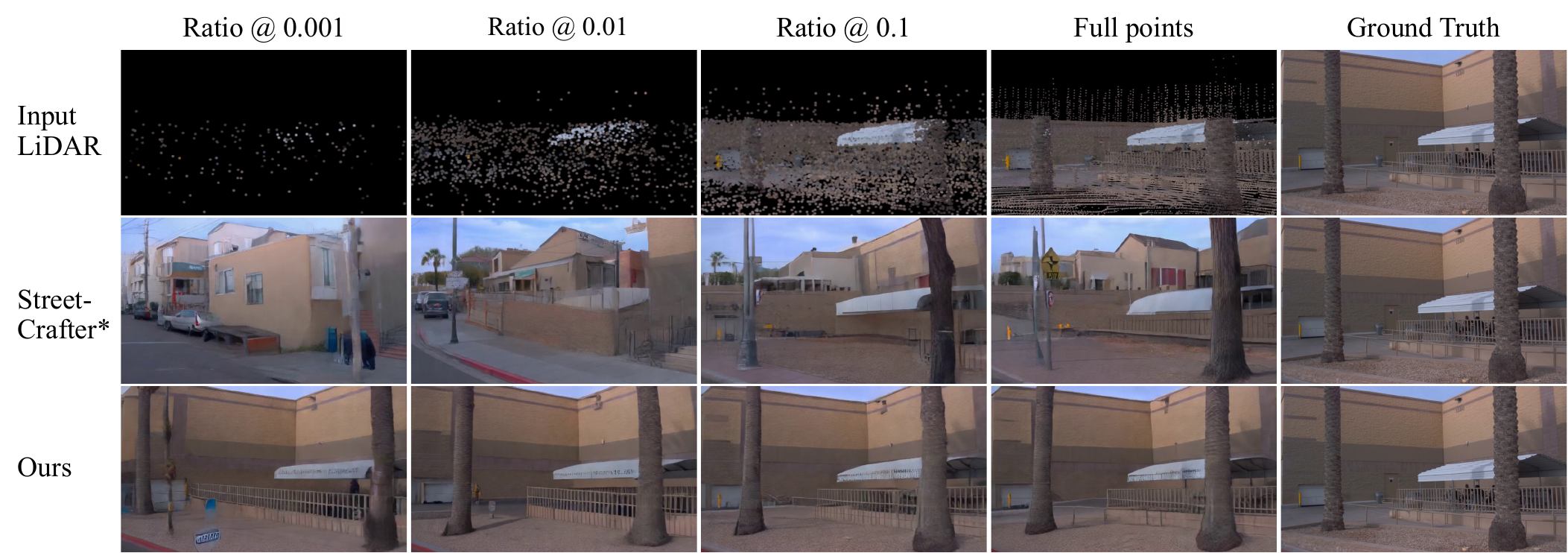}
\caption{
    \textbf{Qualitative comparison across varying sparsities.} Each row shows a single frame from the synthesized video of all methods. Our method preserves high reconstruction fidelity across all levels of LiDAR sparsity, substantially outperforming the baseline.
}
\label{fig:qualitative_multi_ratio}
\end{figure*}
\begin{figure*}[t]
\centering
\includegraphics[width=\linewidth]{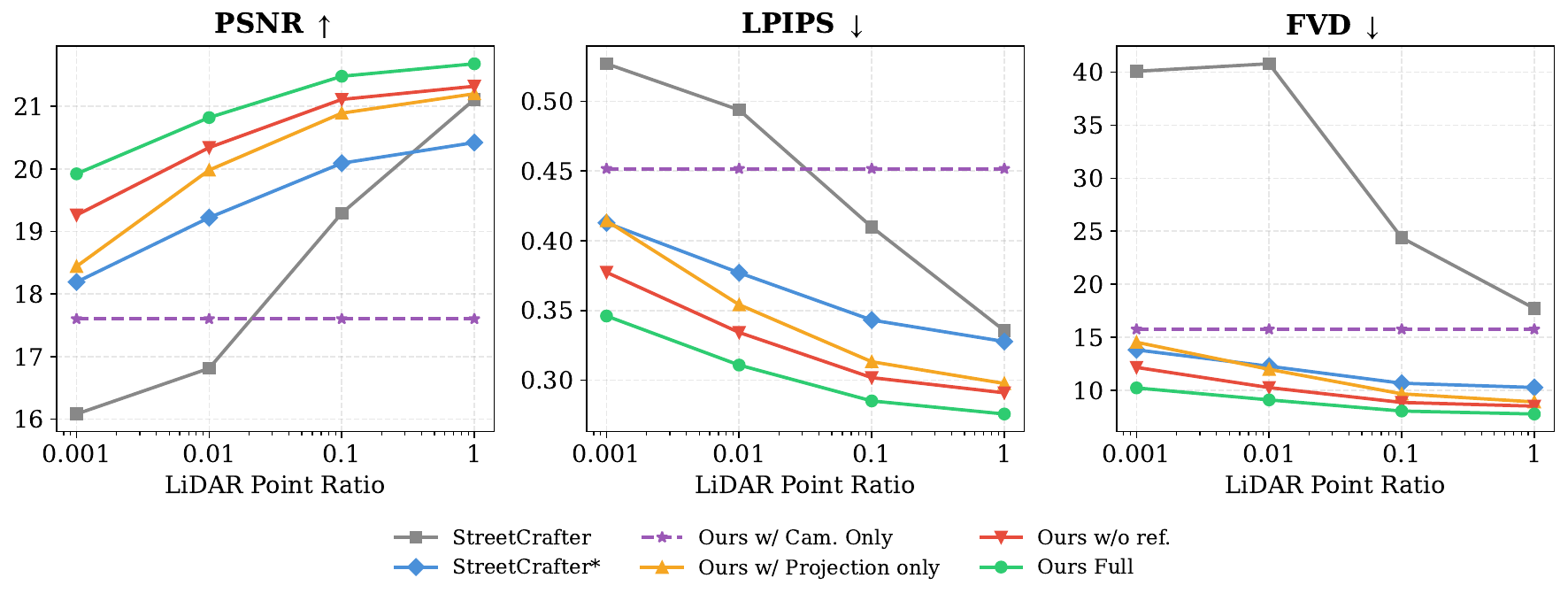}
\caption{
    \textbf{Quantitative Comparison across LiDAR densities.} Our full model consistently outperforms all baselines and ablation variants. The gap narrows as density increases, since dense points gradually dominate the conditioning signal, but our model remains best throughout. Note that ratio $1$ corresponds to full LiDAR and represents the upper bound of density achievable in our benchmark.
}
\label{fig:quantitative_multi_ratio}
\end{figure*}

Beyond the fixed-sparsity setting, we further evaluate across the full range of LiDAR ratios in our dataset, from $0.001$ to $1$.
Figure~\ref{fig:quantitative_multi_ratio} reports our full model, the ablation variants, and both versions of StreetCrafter.
The original StreetCrafter, not exposed to sparse LiDAR during training, degrades sharply as density decreases, while the fine-tuned variant improves in the sparse regime but weakens in the dense one.
Our model outperforms all baselines across the entire spectrum, and the ablation comparison shows that the gains from our camera attention, LiDAR embedding, and reference-view conditioning grow more pronounced as LiDAR becomes sparser, demonstrating the complementary nature of the signals from multi-sensors.

Figure~\ref{fig:qualitative_multi_ratio} provides qualitative comparisons between StreetCrafter* and our method.
At low sparsity ratios, rendered LiDAR points become widely spaced relative to their render radius, producing scattered blobs that no longer convey continuous scene structure; StreetCrafter* fails to recover the correct scene under these conditions.
Our method, in contrast, produces nearly identical reconstructions across all sparsity levels, preserving fine details such as the awning, railing, and palm tree even at the sparsest settings, reflecting how reference views and camera-pose conditioning supply cues that remain stable independent of LiDAR density.

\subsection{Unseen Novel-View Synthesis}

\begin{figure*}[t]
\centering
\includegraphics[width=\linewidth]{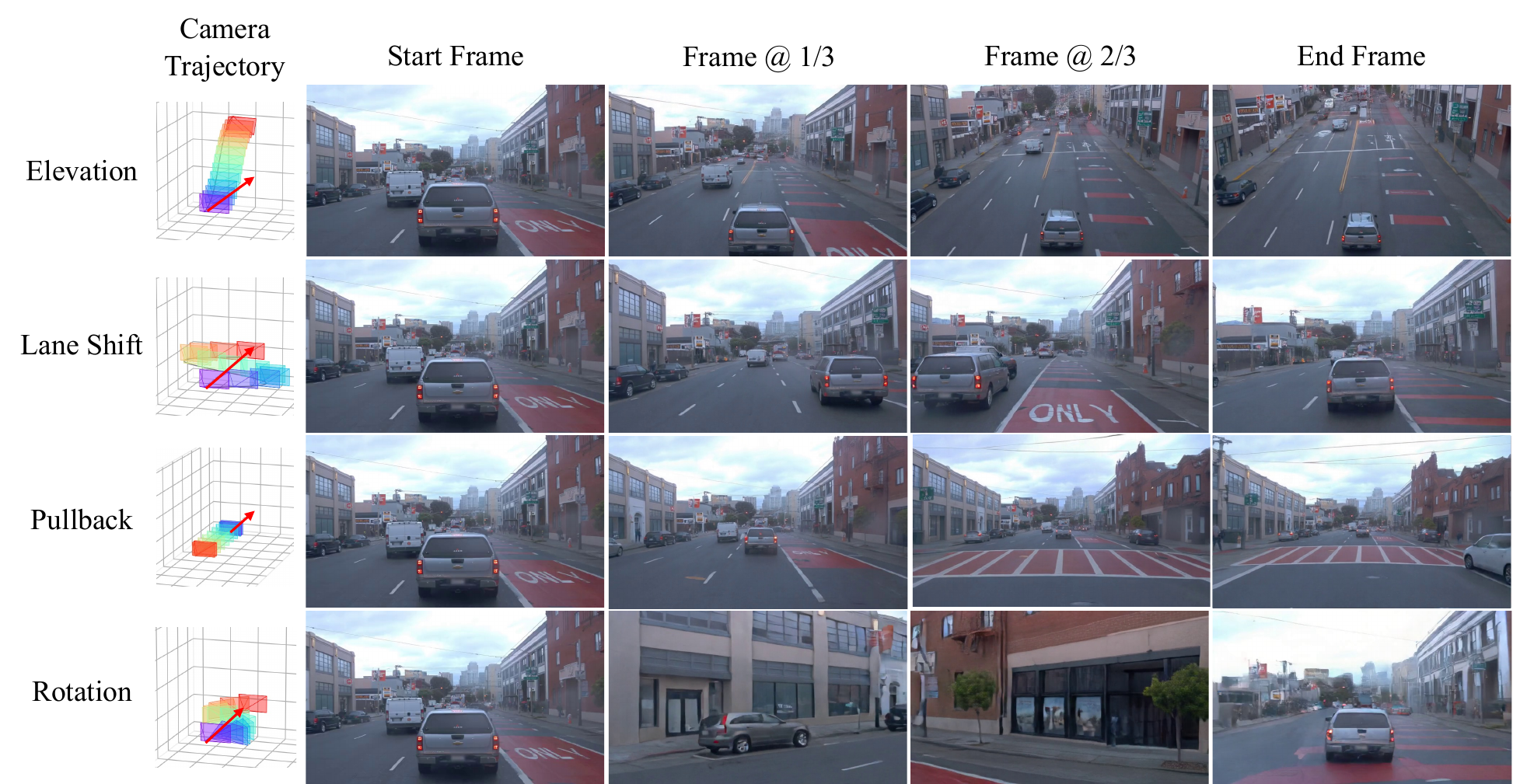}
\looseness=-1
\caption{
    \textbf{Results on unseen trajectories far from the recorded path.} The red arrow in the camera visualization indicates the facing direction of the vehicle. Our method handles a variety of extreme novel trajectories absent from the training data while maintaining high coherence and consistency.
}
\label{fig:demo_novel_view}
\vspace{-10pt}
\end{figure*}

To assess generalization beyond the training distribution, we evaluate our model on four categories of novel-view trajectories not seen during training and far from the recorded driving path: \textit{elevation}, in
which the camera is gradually raised above the vehicle; \textit{lane-shift}, in which the camera is translated to adjacent lanes; \textit{pull-back}, in which the camera dollies straight backward along its forward axis; and \textit{rotation}, in which the camera turns left and right to look around. 

As shown in Figure~\ref{fig:demo_novel_view}, our model preserves scene coherence across all four trajectory types. The rotation case is particularly demanding: the model successfully reconstructs the forward-facing scene at the end of the trajectory to match the starting
frame, despite an extended interval in the middle during which the frontal region is entirely unobserved. The pull-back case similarly stresses geometric and amodal completion: the camera retreats far behind its original position, requiring the model to synthesize plausible content for large disoccluded regions while keeping the originally visible structure stable across frames.



\section{Conclusion}\vspace{-3pt}
\label{sec:conclusion}

\looseness=-1
We presented \method, a unified framework for street-view NVS that jointly conditions a video diffusion backbone on three complementary signals from the vehicle rig: LiDAR reprojections, per-frame camera poses, and surround-view reference imagery.
LiDAR pins down geometry where observed, while pose-aware reference views, fused through our Reference-Enhanced Camera Attention module with relative ray-level positional encoding, fill in appearance and structure elsewhere; a two-stage curriculum further enables robustness across LiDAR sparsity levels at test time.
\method substantially outperforms state-of-the-art baselines under sparse LiDAR, matches methods that rely on $10$--$100\times$ denser point clouds, and generalizes to extreme out-of-trajectory viewpoints such as elevation, lane-shift, spiral, and rotation.
On the other hand, our current evaluation is limited to clips of moderate length, and scaling to longer driving sequences is a natural next direction.
We also focus exclusively on novel-view generation from existing scene content; extending \method toward generative scene editing, such as inserting or removing objects, is a promising direction for future work.

\looseness=-1
\paragraph{Societal Impact.}\vspace{-3pt}
\method is one step toward multi-sensor street-view synthesis as a practical building block for autonomous-driving world models, closed-loop safety simulation, and large-scale data augmentation.
At the same time, like other photorealistic video generators, it could in principle be misused to fabricate misleading driving footage, and synthesized views that look convincing but are subtly geometrically wrong should not be over-trusted in safety-critical pipelines without careful validation.

\paragraph{Acknowledgement.}\vspace{-3pt}
The authors gratefully acknowledge support from Rivian and the Toyota Research Institute (TRI). Some of the computing for this project was performed on the Marlowe~\cite{kapfer_2025_14751899} cluster managed by Stanford Data Science and administered by Stanford Research Computing. We thank the support from NVIDIA Academic Grant Program for this project.

{\small
  \bibliographystyle{ieee}
  \bibliography{main}
}


\clearpage
\appendix

\section{More Results}

Please check our website (\url{https://streetnvs.github.io}) for all animated results.

We provide a finer-grained comparison with baselines by splitting the evaluation data into an \emph{easy} set, consisting of frontal cameras (Camera \#0, \#1, \#2) whose initial views align closely with the target trajectory, and a \emph{hard} set, consisting of side cameras (Camera \#3, \#4) whose initial views overlap only marginally with the rest of the video.
As shown in Table~\ref{tab:supp_quant_comparison_split}, our method outperforms all baselines on both subsets, with a particularly large margin on the hard cases where baselines degrade substantially while our method maintains performance close to that of the easy cases.

\begin{figure*}[ht]
\centering
\includegraphics[width=1.0\linewidth]{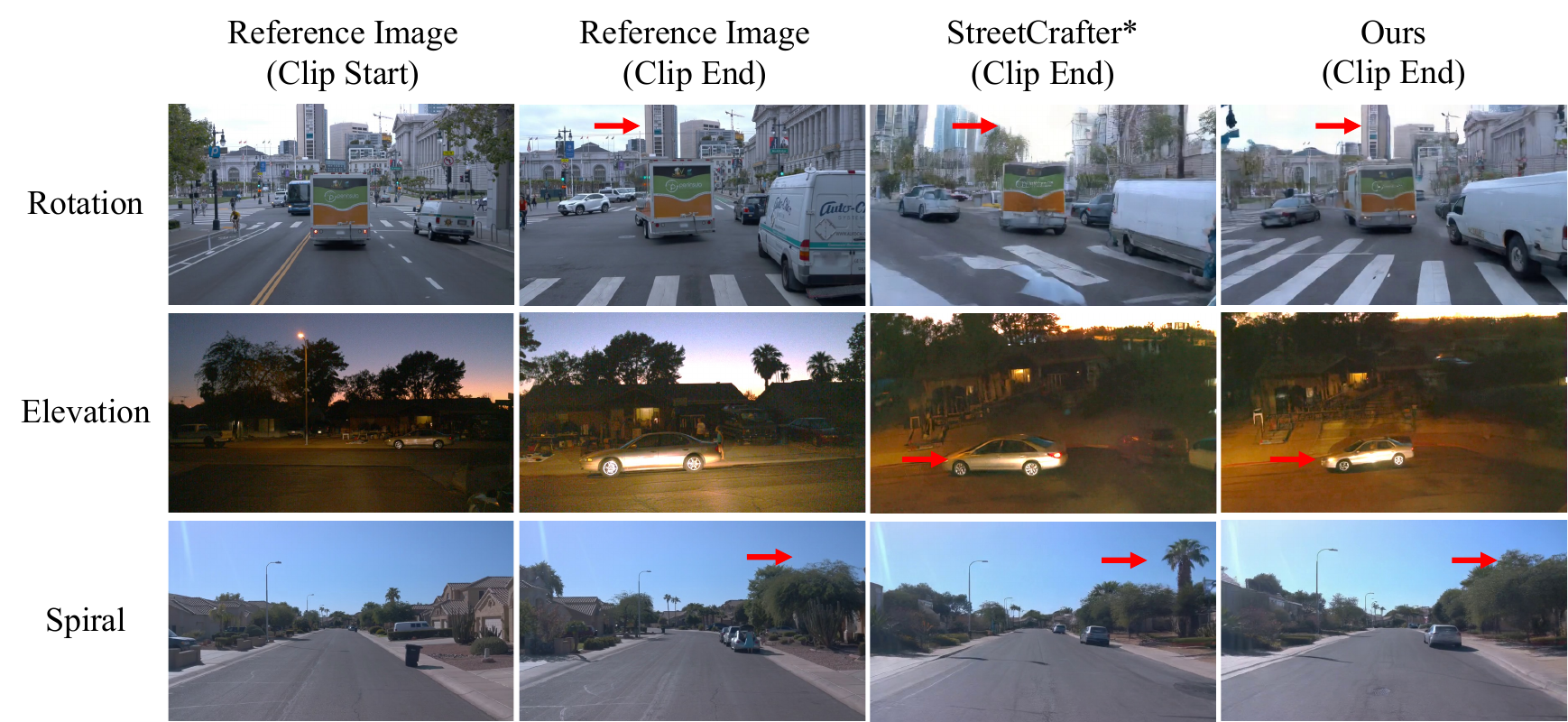}
\caption{
    \textbf{Qualitative Comparison on Unseen Novel Views.}
    For each trajectory, we show the start and end frames of the original vehicle video (left two columns) alongside the end-frame predictions of StreetCrafter$^{*}$ and our model.
    Across rotation, elevation, and spiral cases, our model produces more coherent and geometrically faithful results: it preserves correct background structure under rotation, respects the elevated viewpoint's perspective, and maintains the correct metric scale in the spiral case where the baseline still shows scene content the camera should have already passed.
}
\label{fig:supp_novel_view_comparison}
\end{figure*}

We further compare against StreetCrafter$^{*}$, the strongest baseline in our comparison, on our novel-view trajectories.
As shown in Figure~\ref{fig:supp_novel_view_comparison}, by jointly leveraging reference views and camera-pose conditioning in addition to LiDAR, our model produces results that are more structurally coherent and more geometrically faithful than the baseline across all three trajectory types.

\begin{figure*}[t]
\centering
\includegraphics[width=1.0\linewidth]{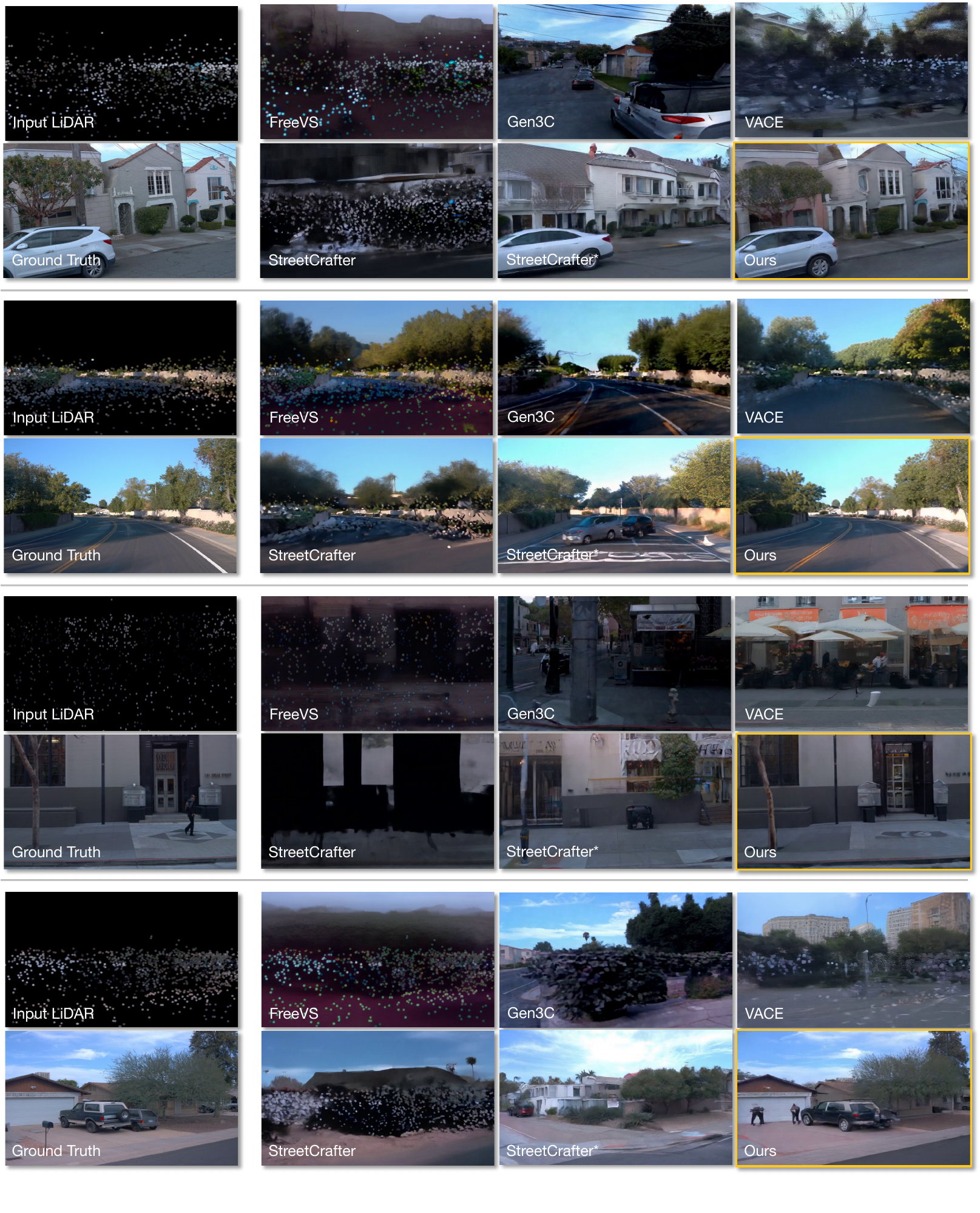}
\vspace{-0.7cm}
\caption{
    \textbf{More Qualitative comparison on the evaluation dataset}. All models are evaluated with 0.01 LiDAR sparsity ratio. Our method substantially outperforms all baselines.
}
\label{fig:supp_qualitative_evaluation}
\end{figure*}

\begin{figure*}[t]
\centering
\includegraphics[width=1.0\linewidth]{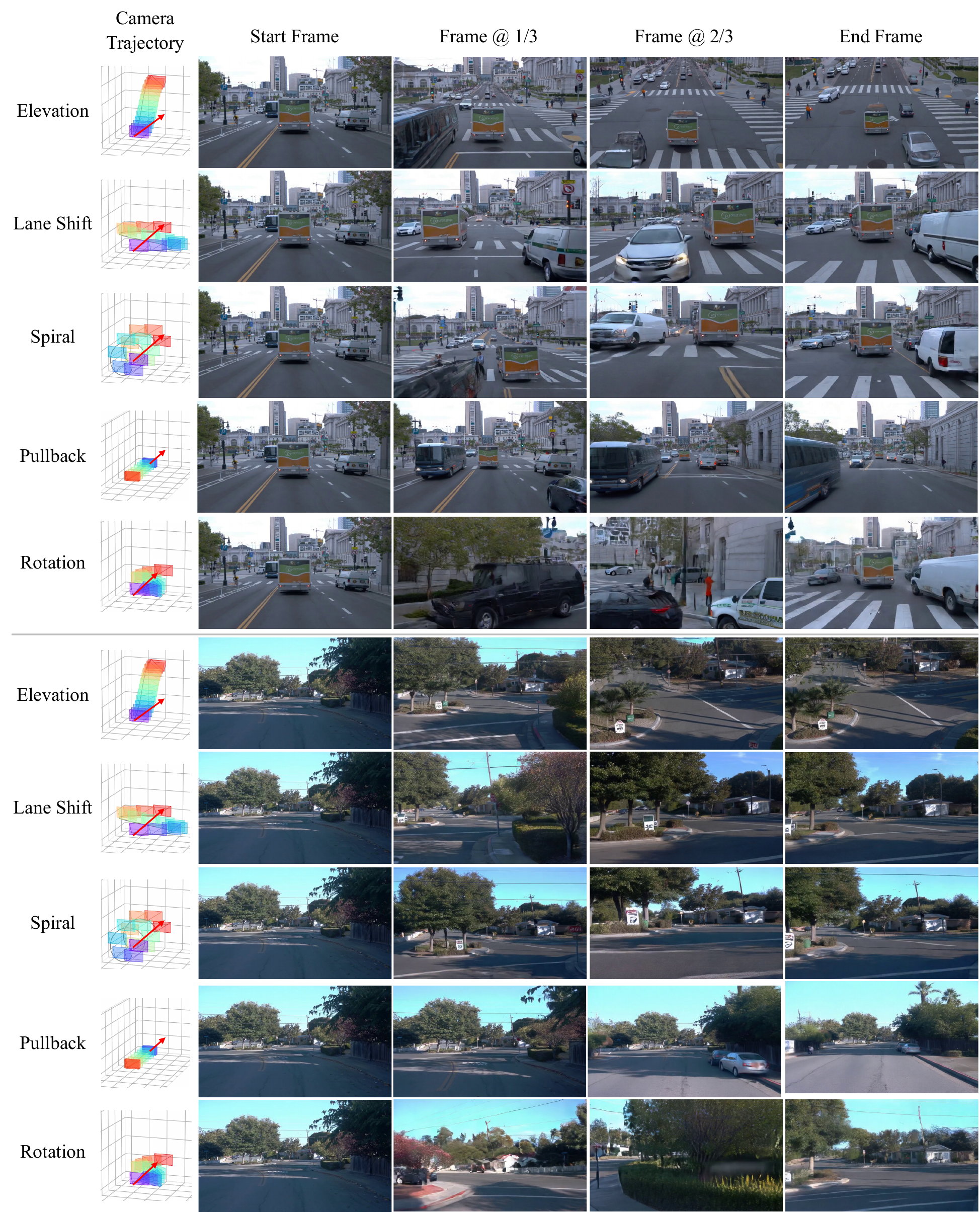}
\vspace{-0.7cm}
\caption{
    \textbf{More results on Unseen Novel Views}.
}
\label{fig:supp_qualitative_novel_view}
\end{figure*}

Beyond the qualitative results in the main paper, we provide additional comparisons in Figure~\ref{fig:supp_qualitative_evaluation} on the evaluation set, and in Figure~\ref{fig:supp_qualitative_novel_view} on the novel-view trajectories.

\begin{table}[h]
    \centering
    \caption{\textbf{Quantitative Comparison by Camera Viewpoint.}
    Evaluated at sparsity ratio $0.01$.
    Frontal cameras ($0/1/2$) provide initial views well-aligned with the target trajectory, while side cameras ($3/4$) provide initial views that overlap only marginally with the rest of the video.}
    \label{tab:supp_quant_comparison_split}
    \setlength{\tabcolsep}{4.5pt}
    \renewcommand{\arraystretch}{1.05}
    \begin{tabular}{l cccc cccc}
        \toprule
        & \multicolumn{4}{c}{\textbf{Frontal Cameras (Easy)}}
        & \multicolumn{4}{c}{\textbf{Side Cameras (Hard)}} \\
        \cmidrule(lr){2-5}\cmidrule(lr){6-9}
        Method
        & PSNR\,$\uparrow$ & SSIM\,$\uparrow$ & LPIPS\,$\downarrow$ & FVD\,$\downarrow$
        & PSNR\,$\uparrow$ & SSIM\,$\uparrow$ & LPIPS\,$\downarrow$ & FVD\,$\downarrow$ \\
        \midrule
        FreeVS~\cite{wang2024freevs}              & 12.51 & 0.4116 & 0.7245 & 197.93 & 13.87 & 0.4030 & 0.7172 & 179.41 \\
        Gen3C~\cite{ren2025gen3c}                 & 14.49 & 0.4517 & 0.5487 &  38.49 & 14.49 & 0.4028 & 0.6542 &  45.61 \\
        VACE~\cite{vace}                          & 15.27 & 0.4726 & 0.5548 &  78.69 & 14.07 & 0.3880 & 0.6914 &  77.52 \\
        StreetCrafter~\cite{yan2024streetcrafter} & 17.92 & 0.5636 & 0.4150 &  43.48 & 15.70 & 0.4211 & 0.5726 &  61.47 \\
        StreetCrafter$^{*}$                       & 19.65 & 0.6065 & 0.3186 &  15.31 & 18.80 & 0.5584 & 0.4354 &  23.31 \\
        \midrule
        \textbf{StreetNVS (Ours)}                 & \textbf{20.93} & \textbf{0.6364} & \textbf{0.2710} & \textbf{11.60}
                                                  & \textbf{20.71} & \textbf{0.6018} & \textbf{0.3506} & \textbf{16.27} \\
        \bottomrule
    \end{tabular}
\end{table}

\section{Additional Implementation Details}

\subsection{Model}

\paragraph{\textbf{Backbone and Module Integration.}}
Our model is built on top of the WAN-2.2-I2V-5B~\cite{wan2025wan} backbone.
Aside from the original model, the LiDAR embedding layer consists of two convolutional layers followed by a two-layer MLP that projects the LiDAR features into the DiT's hidden dimension, and the Camera Attention branch is added in parallel within each DiT block as described in Section~\ref{sec:method}.
To accommodate reference inputs, we encode reference views frame-by-frame with the WAN VAE encoder, producing the same frame count in token space.
To disambiguate reference tokens from target tokens within the noisy latent, we offset their temporal index by a constant of $64$, well beyond the maximum temporal length of the latent space in our setup, $\lfloor 81/4 \rfloor = 20$.
The reference tokens are then padded with constant LiDAR placeholder channels and concatenated with the target tokens, after which they are treated identically to target tokens by the network.

\paragraph{\textbf{Camera Pose Encoding.}}
We use the registered camera poses provided with the dataset.
Because the video VAE compresses several frames into a single token along the temporal axis, we use the pose of the first frame within each compressed group as the representative pose for the corresponding token.
Empirically, alternative choices (e.g., the middle or last frame) yield similar results.
Since UCPE~\cite{zhang2025unified} is scale-sensitive, we rescale the scene by a factor of $0.1$ relative to the metric scale to match the regime in which UCPE performs best.
In all experiments, the camera-pose encoding occupies $3/4$ of each attention head's channels ($96$ channels), while the remaining $1/4$ ($32$ channels) is reserved for the frame-index encoding.

\subsection{Data}

\paragraph{\textbf{Data Processing.}}
For each scene in the Waymo Open Dataset, we sample LiDAR scans at every $10$th frame (i.e., frames $0, 10, 20, \dots$) and aggregate them into a single point cloud, which we then uniformly subsample at several sparsity ratios.
The aggregated points are rendered into the target trajectory using the differentiable point renderer of StreetCrafter~\cite{yan2024streetcrafter} with a point radius of $0.02$ in NDC space.
Rendered depth maps are normalized to $[0, 1]$ and converted to RGB via a perceptual colormap (Jet).
Training clips are extracted as $49$-frame subsequences with a stride of $10$ frames, yielding approximately $61{,}000$ clips per sparsity ratio.
The evaluation set is constructed analogously, but uses only the first $49$ frames of each scene, producing $402$ evaluation clips in total.

\paragraph{\textbf{Novel-View Trajectory Construction.}}
To evaluate generation under unseen viewpoints, we construct five families of novel camera trajectories that deviate from the original driving path.
Let $f \in \{0, \dots, F-1\}$ index frames, $t = f/(F-1) \in [0, 1]$ denote normalized time, and $\bm{T}_f$ the original front-camera-to-world pose.
We decompose $\bm{T}_f$ as
\begin{equation*}
    \bm{T}_f = \begin{bmatrix} \bm{R}_f & \mathbf{c}_f \\ \mathbf{0} & 1 \end{bmatrix}
    = \big[\mathbf{r}_f \mid \mathbf{d}_f \mid \mathbf{f}_f \mid \mathbf{c}_f\big],
\end{equation*}
where $\bm{R}_f = [\mathbf{r}_f, \mathbf{d}_f, \mathbf{f}_f]$ is the rotation matrix whose columns are the camera's right, down, and forward axes, and $\mathbf{c}_f$ is the camera center.
The four trajectory families are defined as follows.

\noindent\textit{\textbf{Elevation.}}
The camera is gradually lifted toward a bird's-eye view by interpolating between the original pose and an elevated, pitched target pose:
\begin{equation*}
    \bm{T}_f^{\mathrm{ele}}(t) = \operatorname{Lerp}(\bm{T}_f, \bm{T}_f^{\mathrm{BEV}}, t),
\end{equation*}
where $\bm{T}_f^{\mathrm{BEV}}$ is constructed by lifting the camera to height $h = 5$\,m, shifting it backward by $b = 2$\,m, and applying a downward pitch of $\theta = 30^\circ$.

\noindent\textit{\textbf{Spiral.}}
The camera orbits the original viewpoint along a circular path of radius $r = 1.5$\,m, completing $K = 2$ full revolutions while keeping its orientation fixed:
\begin{equation*}
    \mathbf{c}_f^{\mathrm{spi}}(t) = \mathbf{c}_f + r\big[(\cos\alpha - 1)\,\mathbf{r}_f + \sin\alpha\,\mathbf{d}_f\big],
    \quad \alpha = 2\pi K t, \quad \bm{R}_f^{\mathrm{spi}} = \bm{R}_f.
\end{equation*}

\noindent\textit{\textbf{Lane-Shift.}}
The camera is laterally displaced by a sinusoidal offset of amplitude $A = 3.5$\,m along the lateral axis $\hat{\mathbf{r}}_f$ (orthogonal to the driving direction), simulating lane-changing behavior.
\begin{equation*}
    \mathbf{c}_f^{\mathrm{lane}}(t) = \mathbf{c}_f + s A \sin(2\pi t)\,\hat{\mathbf{r}}_f,
    \quad \bm{R}_f^{\mathrm{lane}} = \bm{R}_f,
\end{equation*}
\looseness=-1
where $s \in \{-1, +1\}$ is a per-scene sign that determines whether the camera shifts to the left or right adjacent lane.

\noindent\textit{\textbf{Rotation.}}
The camera yaw oscillates within a range of $\psi_{\max} = 60^\circ$, smoothly interpolating between a set of precomputed camera-to-world poses $\{\bm{T}_f^{(j)}\}_{j=0}^{4}$ corresponding to canonical yaw angles $\psi^{(j)} \in [-90, -45, 0, 45, 90]^\circ$:
\begin{equation*}
    \psi(t) = \psi_{\max}\sin(2\pi t), \quad
    \bm{T}_f^{\mathrm{rot}}(t) = \operatorname{Lerp}(\bm{T}_f^{(j)}, \bm{T}_f^{(j+1)}, \beta),
\end{equation*}
where $j$ is the index such that $\psi^{(j)} \le \psi(t) < \psi^{(j+1)}$, and $\beta = (\psi(t) - \psi^{(j)}) / (\psi^{(j+1)} - \psi^{(j)}) \in [0, 1]$ is the linear interpolation weight.
The wider precomputed pose set supports flexible yaw selection; in all our experiments, the actual oscillation stays within $\psi_{\max} = 60^\circ$.

In all uses of $\operatorname{Lerp}(\cdot)$, the rotation component of the interpolated pose is re-orthonormalized to ensure the result lies in $\mathrm{SE}(3)$.

\subsection{Training}

We fine-tune the WAN backbone end-to-end together with the newly introduced modules.
Since fine-tuning all parameters of the original model already requires $\sim\!70$\,GB of GPU memory, we apply LoRA with rank $1024$ to the FFN layers, reducing memory consumption to $\sim\!40$\,GB and leaving sufficient headroom for the additional modules introduced by our framework.

We adopt a two-stage progressive training curriculum.
In the first stage, we sample data from the four sparsity ratios (from $0.001$ to $1$) with weights $\{0, 0.25, 0.25, 0.5\}$, disable the Camera Attention branch, and train only the original WAN backbone together with the LiDAR embedding layers, using the rendered LiDAR buffers as the sole conditioning signal.
In the second stage, we rebalance the sparsity distribution to $\{0.15, 0.35, 0.35, 0.15\}$ and fine-tune the full model with all modules enabled.
Both stages run for $10$K iterations under the same learning-rate schedule.



\end{document}